\documentclass[runningheads,a4paper]{llncs}

\usepackage{amssymb}
\usepackage{color}
\usepackage{tabularx}

\usepackage{graphicx}

\newcommand{\figref}[1]{Fig.~\ref{#1}}
\newcommand{\tabref}[1]{Tab.~\ref{#1}}

\newcommand{\eqref}[1]{Eq.~\ref{#1}}
\usepackage{bbding}

\usepackage{url}
\urldef{\mailsa}\path|{zhaobinnku}@mail.nankai.edu.cn|
\newcommand{\keywords}[1]{\par\addvspace\baselineskip
\noindent\keywordname\enspace\ignorespaces#1}

\graphicspath{{./figure/}}

\begin{document}

\mainmatter  

\title{A Multi-Stage Framework for 3D Individual Tooth Segmentation in Dental CBCT }

\titlerunning{Lecture Notes in Computer Science: Authors' Instructions}

%
%
\author{Chunshi Wang \inst{1}%
\and   Bin Zhao\inst{1,2}\textsuperscript({\Envelope}\textsuperscript)  \and   Shuxue Ding\inst{1,2}
}
\authorrunning{Lecture Notes in Computer Science: Authors' Instructions}

\institute{School of Artificial Intelligence, Guilin University of Electronic Technology, Guilin, Guangxi, 541004, China \\ \mailsa\\ \and Guangxi Colleges and Universities Key Laboratory of AI Algorithm Engineering, Guilin, Guangxi, 541004, China  
}

%
%

\toctitle{Lecture Notes in Computer Science}
\tocauthor{Authors' Instructions}
\maketitle

\begin{abstract}
Cone beam computed tomography (CBCT) is a common way of diagnosing  dental related diseases. Accurate segmentation of 3D tooth is of importance for the  treatment. Although deep learning based methods have achieved convincing results in medical image processing, they need a large of annotated data for network training, making it very
time-consuming in data collection and annotation. Besides, domain shift widely existing  in the distribution of data acquired by different devices impacts severely the model generalization. To resolve the problem, we propose a multi-stage framework for 3D  tooth segmentation in dental CBCT, which achieves the third place in the "Semi-supervised Teeth Segmentation" 3D (STS-3D) challenge. The experiments on validation set compared with other semi-supervised segmentation methods further indicate the validity of our approach.
\end{abstract}

\section{Introduction}
Advances in modern digital dentistry rely heavily on the acquisition and segmentation of three-dimensional (3D) imaging. In particular, cone beam computed tomography (CBCT) plays a crucial role in obtaining accurate 3D digital models of the jaws and teeth while keeping costs and radiation doses relatively low.

The acquisition and segmentation of 3D digital images has a variety of applications in the field of oral and maxillofacial disciplines, with the most prominent applications being orthodontic diagnosis and treatment planning. Specifically, these techniques are commonly used in 3D-guided implant surgery, guided endodontic and apical surgery, CBCT-based planning and fabrication of donor tooth replicas, etc. 3D tooth segmentation is an important part of the aforementioned treatment processes and can be used in orthodontics to develop treatment plans and to follow the evolution of root resorption after treatment. 
However, precise tooth segmentation is challenging. The presence of a large number of teeth on each arch in the mouth  complicates the segmentation process. Teeth in the emergence stage have some specialized structures that make them difficult to distinguish. Metal fillings and dental restorations can cause artifacts in CBCT, which can lead to aberrations in the segmentation results. In addition, the composition of the tooth itself is more complex, with elements consisting of cementum, dentin, pulp  and enamel in close contact with other anatomical structures in the mouth, making it difficult to determine tooth edges. There is also a similar density between the alveolar bone and the tooth structure, which leads   to separate the upper and lower teeth from each other difficultly \cite{jang2021fully}.
\vspace{-1pt}

Over the past decade, many attempts have been made to develop 3D tooth segmentation methods, most of which have been level-set based methods \cite{gao2010individual,ji2014level}. Unfortunately, these methods do not achieve fully automatic segmentation for the reason that they need to manually intervene in the initialization of the level-set and the complex image structure between adjacent teeth and bones further prevents automatic initialization. Besides, level-set based methods require a lot of mathematical operations and perform poorly for metal artifacts and upper-lower tooth separation. There are also some  graph cut based methods for tooth segmentation \cite{hiew2010tooth,evain2017semi}, but they need  priors to guide the segmentation process and  lack robustness.

Recently, convolutional neural networks (CNNs) have been applied to 3D tooth segmentation aiming to overcome the limitations of traditional segmentation methods. The focus of researchers has shifted to the development of an algorithm for fully automated tooth segmentation without human intervention, striving for accuracy and speed \cite{polizzi2023tooth}. For instance, Chen et al. propose a 3D full convolutional neural network (FCN) combined with the watershed transform for tooth segmentation \cite{chen2020automatic}. Hsuet al. propose a 3.5D U-Net model to improve the performance of tooth segmentation \cite{hsu2022improving}. Cui et al. propose Toothnet, a network that utilizes edge maps, similarity matrix  and spatial relationships between teeth \cite{cui2019toothnet}. Cui et al. further develop an AI system for fully automated segmentation of teeth and alveolar bone using the 2-Stage scheme, which is validated on a large number of CBCT scans, demonstrating to some extent the effectiveness of AI technology in enhancing clinical workflow \cite{cui2022fully}.

The above-mentioned methods are performed in fully supervised scenarios, where the data need to be labeled in detail by experts. However, the labeling process for 3D medical data is both complex and expensive. In addition, there are limited datasets available for 3D tooth segmentation studies \cite{cui2022ctooth}, which severely limits its wide application in the field of tooth segmentation. To address this problem, semi-supervised learning (SSL) methods for medical image segmentation have emerged \cite{zhao14automatic,zhao2022combine}. These methods require less expert annotation for model training, thus reducing the time and effort required for data annotation. But Since domain offsets exist widely in medical data, semi-supervised learning cannot directly obtain better segmentation results. Therefore, in this paper, we propose a multi-stage framework based on SSL and domain adaptation to segment tooth from CBCT, which achieves the third place in the "Semi-supervised Teeth Segmentation" 3D (STS-3D) challenge. The  extensive experiments  compared with other semi-supervised segmentation methods further indicate the validity of our approach.

\section{Materials and Method}

\subsection{Research subjects}
The experimental data used in this paper is provided by the STS-3D challenge where training set includes 12 CT scans with annotations and 300 CT scans without annotations \cite{cui2022ctooth,cui2022ctooth+}. There are 50 CT scans for test and they do not provided the annotations for participants.

\subsection{Multi-stage framework}
\begin{figure}[h]
	\centering
	\setlength{\abovecaptionskip}{0.3cm}
	\includegraphics[width=\linewidth]{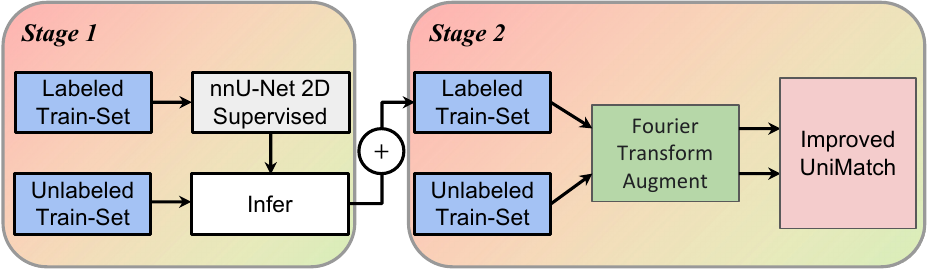} \\
	\caption{The architecture of our proposed multi-stage framework. Best viewed in color.}
	\label{fig:framework}
\end{figure}

Our proposed multi-stage framework is illustrated in the \figref{fig:framework}. As shown in the \figref{fig:framework}, in stage 1, we utilize supervised learning based on the 2D nnU-Net\cite{isensee2021nnu} model to generate pixel-level pseudo-annotations for a small number of randomly sampled unlabeled samples from the training set, and then combine these pseudo-annotated  samples with the labeled samples in the training set to form new labeled samples. In stage 2, these new labeled samples are fed into the Improved-UniMatch model for parameter learning after domain adaptation together with the unlabeled samples in the training set.

In particular, in stage 1, the labeled samples is first used to train the 2D nnU-Net 20 epochs, and then the trained 2D nnU-Net  is used to generate low-quality pixel-level pseudo-annotations for the 10 randomly selected unlabeled samples in the unlabeled set. These 10 pseudo-annotated samples are then merged with the labeled samples to  form new labeled samples to train the stage 2. 
\begin{figure}[h]
	\centering
	\setlength{\abovecaptionskip}{0.3cm}
	\includegraphics[width=\linewidth]{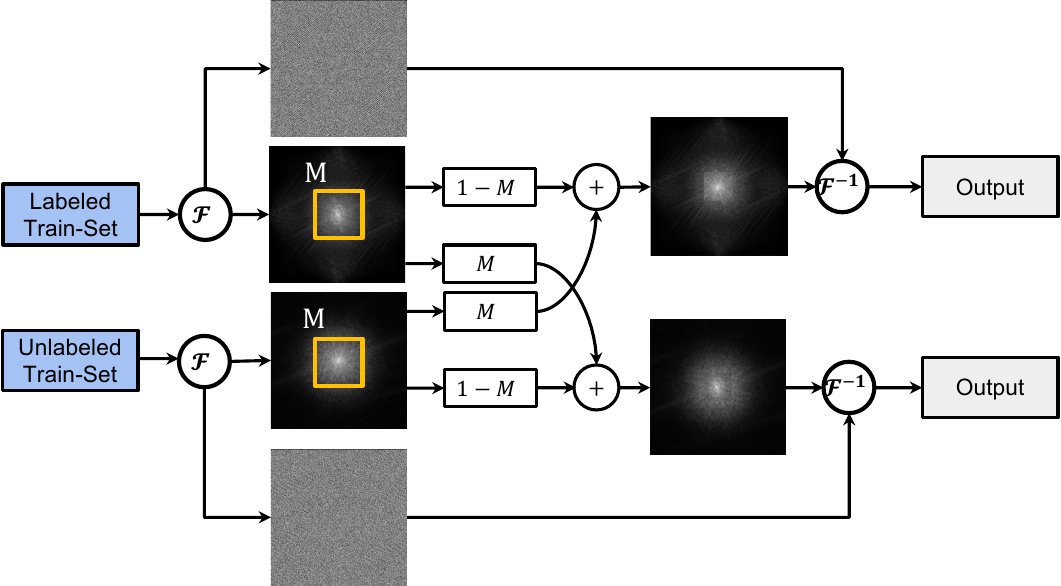} \\
	\caption{The flowchart of FTA. }
	\label{fig:FTA}
\end{figure}
In stage 2, the new labeled samples and the remaining unlabeled samples are fed into Fourier Transform Augment (FTA) module to make the model learn the difference between two domains. The FTA module is presented in \figref{fig:FTA}. During the training process, a labeled image $x^w$ and a unlabeled image $x^u$ are randomly selected, and then a Fourier transform $\mathcal{F}$ is performed to transfer the images to the frequency domain and obtain the magnitude spectrum \{$A^w$, $A^u$\} and the phase image \{$P^w$, $P^u$\}, where the magnitude spectrum contains the low-level statistics and the phase image includes the high-level semantics of the original signal \cite{yao2022enhancing}. The image $x^w$ is then enhanced by merging the magnitude information of the image $x^u$,
\begin{equation}
    A_{new}^w=(1-\lambda) A^w\ast(1-M)+\lambda A^u \ast  M, 
\end{equation}	
where $A_{new}^w$ is the newly generated phase map. $\lambda$ is a parameter to adjust the ratio between the phase information of $x^w$ and $x^u$ . $M$ is used to control the spatial extent of the magnitude spectrum to be exchanged, and here $M$ is set to the center region of the magnitude spectrum containing low frequency information. After that, the merged samples are transformed from the frequency domain to the image domain by $\mathcal{F}^{-1}$ to obtain the image sample $Z^w$ enhanced by Fourier transform and fused with the low-level information of the other sample,

\begin{equation}
    	Z^w=\mathcal{F}^{-1}(A_{new}^w,P^w).
\end{equation}

Similarly exchanging $x^w$ and $x^u$ gives $Z^u$:

 \begin{equation}
 	Z^u=\mathcal{F}^{-1}(A_{new}^u,P^u).
\end{equation}

It is worth mentioning that $Z^w$ and $Z^u$ can be merged together for computation, which will greatly reduce the computational cost. That is, it is necessary to augment both $x^u$ and $x^w$  at the same time, and send the result of mutual augmentation into the subsequent training process.

The data performed through FTA are fed into the Improved-Unimatch for segmentation training process.  In particular, we introduce the idea of self-adaptive thresholding in FreeMatch \cite{wang2022freematch} into the current segmentation task, allowing UniMatch\cite{yang2023revisiting} to adaptive adjust the threshold during training and improve the segmentation accuracy.
Besides, we improve the feature perturbation to enable better consistency learning of the model by replacing the perturbation of Dropout with an Alpha Dropout with self-normalization \cite{klambauer2017self}.

\subsection{Evaluation metrics}

In this research, 
 dice coefficient  and intersection over union (IoU) are used to evaluate the pixel-level
segmentation performance,  which are formulated as 
\begin{equation}
    Dice = \frac{2\ast|A\cap B|}{|A|+|B|},
\end{equation}
 and
 \begin{equation}
    IoU = \frac{A\cap B}{A \cup B},
\end{equation}
respectively, where A and B indicate the predicted segmentation  and the ground truth, respectively. In addition, the 3D Hausdorff distance is used to evaluate the voxel-level segmentation performance and is formulated as 
\begin{equation}
    H(d) = min(|x1-x2|+|y1-y2|+|z1-z2|),
\end{equation}
where $(x1,y1,z1)$ and $(x2,y2,z2)$ denote the coordinates of the two voxels. $|x1 - x2|$, $|y1 - y2|$ and $|z1 - z2|$ denote the distances on the corresponding axes. 

For scoring purposes, the challenge uniformly normalizes the hausdorff distance to the range of [0, 1]. The final weighted average of the three metrics is taken  and the specific scoring formula is
\begin{equation}
    score = 0.4\ast Dice+0.3\ast IoU+0.3\ast (1-H(d)).
\end{equation}

\section{Experiment}
\subsection{Data preprocessing}

To take advantage of the pixel-level pseudo-annotations generated from unlabeled data and obtain rich tooth morphology, the training data are sliced in 3-axis. That is, the 3D tooth data are sliced in axial, coronal, and sagittal planes, e.g., a 640 $\times$ 640 $\times$ 400 CT image would result in 640 $+$ 640 $+$ 400 $=$ 1680 slices. Using file suffixes as the division of slice axes, for simplicity, $\_x$,$ \_y$ and$ \_z$ are directly used as the suffixes of slice files \cite{liu2018towards}.
In addition, the slices are normalized to a range of [0,1] and 10\% of slices are used as the validation set.

\begin{table}[ht]
	\centering
 \caption{Quantitative  results  of input images with different thresholds on the validation set}
	\begin{tabularx}{\textwidth}{XXXXX} \hline
Bottom  & Top   & val1 & val2  & val3  \\ \hline
	0 & 1500  & 0.774 & 0.818 & 0.862   \\
 	500 & 1500  & 0.817 & 0.826 & 0.876   \\
   	500 & 1800  & 0.839 & 0.804 & 0.885   \\
       550 & 1950  & 0.725& 0.850 & 0.873   \\
    500 & 2000  & \textcolor{red}{0.816}& \textcolor{red}{0.848} & \textcolor{red}{0.884}   \\
        300 & 2000  & 0.813& 0.854 & 0.879   \\
        450 & 2050  & 0.829& 0.820 & 0.867   \\
         400 & 2100  & 0.747& 0.813 & 0.845   \\
                  0 & 2500  & 0.825& 0.830 & 0.860   \\
		\hline
	\end{tabularx}
 
	\label{tab:table_te}
	\vspace{0 in}
\end{table}

In order to explore the effect of different thresholds of the input image on the performance of the proposed method, we conduct thresholding experiments to select the best threshold for training. In particular, to save time, we randomly select 25\% of  slices from the training slices to train our model. We train 3 epochs and the quantitative  results on validation set of each epoch are shown in \tabref{tab:table_te}. From \tabref{tab:table_te}, it can be seen that when the combination of 500 and 2000 is chosen for the threshold value, the proposed method has the best  segmentation results on the validation set, so this combination is chosen as the threshold value in the experiment.

\subsection{Implementation Details}  \label{sec:results}

The optimizer is the AdamW method \cite{loshchilov2017decoupled} with $weight_{decay}=0.0001$. The initial learning rate is $10^{-4}$. During training, the learning rate is updated using the following formula,
\begin{equation}
    lr={lr}_b \times (1- \frac{i}{N})^{0.9},
\end{equation}
where ${lr}_b$ indicates  the initial learning rate. $i$ and $N$ indicate the  current number  and  the total number of iterations, respectively.

The experiments are performed on a computer with an Intel Core i7-6800K CPU, 64 GB RAM and NVIDIA Tesla V100 GPU with 32 GB memory. 
All networks are implemented in PyTorch.

\begin{table*}[h]
	\centering
	\caption{The top five submission results of the STS-3D challenge. Our result has been highlighted in red. }
	\label{tab:table_cha}
	\begin{tabularx}{\textwidth}{XXXXX} \hline
		participant & Score($\uparrow$)  & Dice & IoU & H(d) \\  		\hline
		RoboSurge & 0.8497  & 0.8442 & 0.8661 & 0.1595 \\ 
  		Yxx & 0.8427  & 0.8343 & 0.8583 & 0.1615 \\ 
	\textcolor{red}{GUET-IICI} & \textcolor{red}{0.8256}  & \textcolor{red}{0.8058} & \textcolor{red}{0.8376} & \textcolor{red}{0.1599} \\ 
	IGIP-CBCT & 0.8237  & 0.8070 & 0.8386 & 0.1689 \\ 
 	sdkxd & 0.8147  & 0.7932 & 0.8290 & 0.1708 \\ 
		\hline
	\end{tabularx}
\end{table*}

\subsection{Results} \label{sec:results}

\begin{figure*}[ht]
	\centering
	\vspace{-0.5cm} 
	\setlength{\abovecaptionskip}{-0.1cm}
	\includegraphics[width=\linewidth]{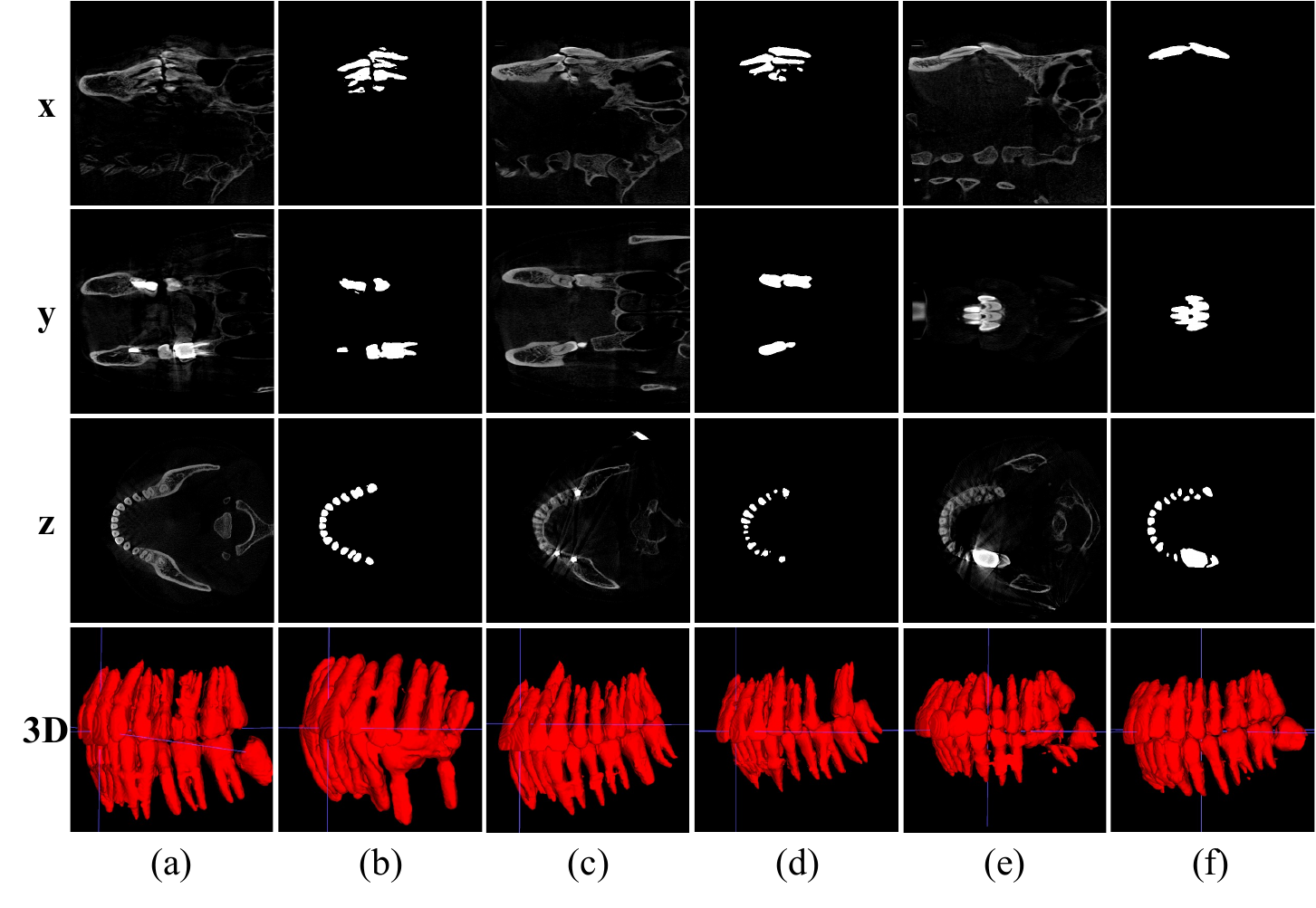} \\
	\caption{Six examples showing segmentation results obtained from our method. The four rows from top to bottom denotes axial, coronal, sagittal and 3D views. Best
viewed in color.}
	\label{fig:visualization_examples}
\end{figure*}

We train our model only 3 epochs for limited time and submit the results to the challenge. \tabref{tab:table_cha} summaries the top five submission results of the STS-3D challenge and our result has been highlighted in red. As shown in the \tabref{tab:table_cha}, our proposed method achieves the third place in the challenge.  In addition, we visualize some segmentation results in the \figref{fig:visualization_examples}. No matter from the  axial, coronal and  sagittal, our proposed method obtains the better segmentation results. 

\begin{figure*}[ht]
	\centering
	\vspace{-0.5cm} 
	\setlength{\abovecaptionskip}{-0.1cm}
	\includegraphics[width=\linewidth]{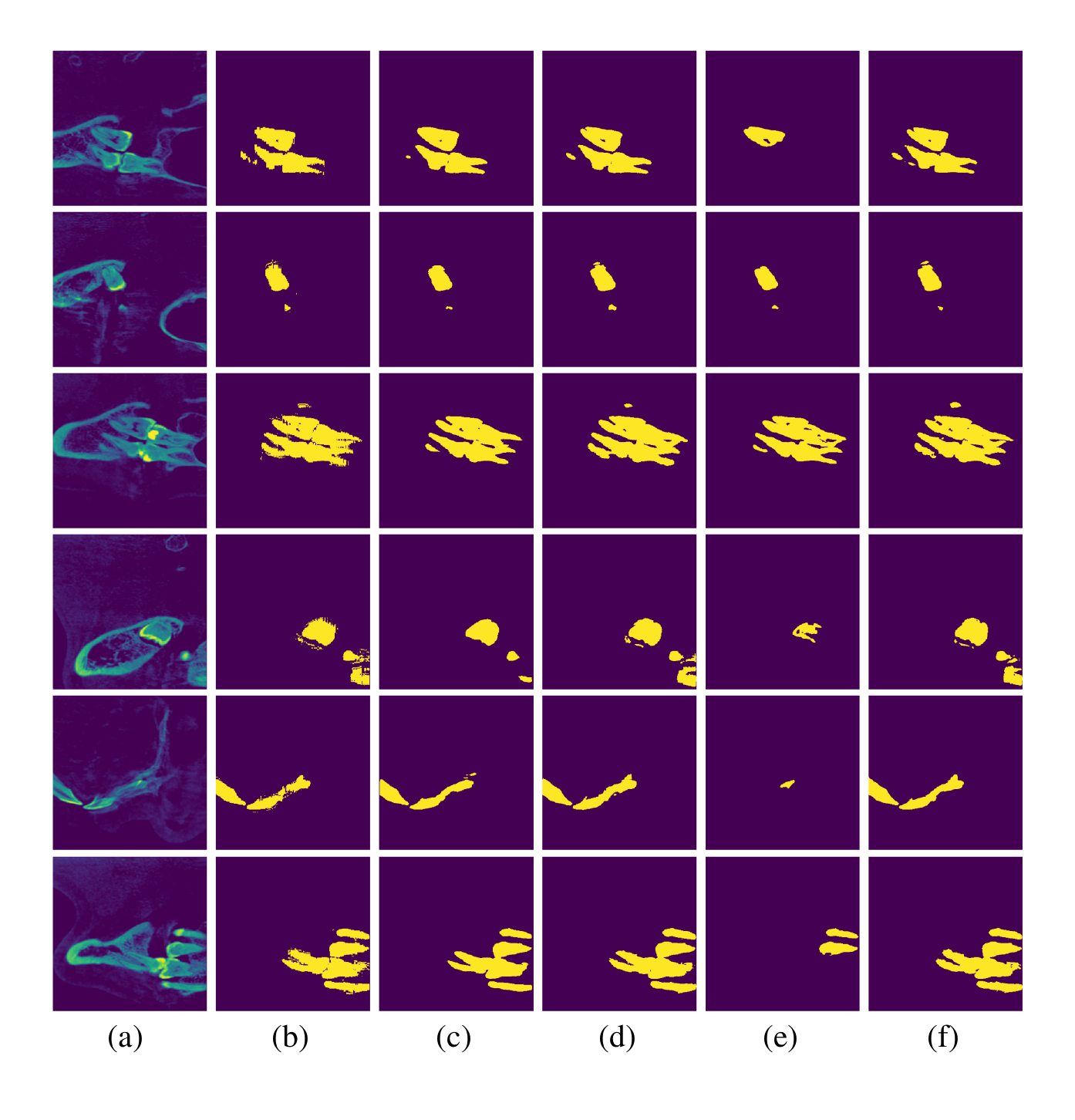} \\
	\caption{Comparison of segmentation performance. (a-b) CBCT slices and their corresponding ground truth. (c-f) are the segmentation results of CPS \cite{chen2021semi},  FixMatch \cite{sohn2020fixmatch}, MeanTeacher \cite{tarvainen2017mean} and our proposed method, respectively.}
	\label{fig:comparison_withsota}
\end{figure*}

We further compared with other state-of-the-art (SOTA) semi-supervised segmentation methods using the challenge data. Specifically, we  train and evaluate  CPS \cite{chen2021semi},  FixMatch \cite{sohn2020fixmatch} and MeanTeacher \cite{tarvainen2017mean} using the same settings on the challenge data. In particular, we use the validation set as the test set to validate the proposed method and the comparison methods.

\figref{fig:comparison_withsota} visualizes some slices of segmentation results and the ground truth. As the \figref{fig:comparison_withsota} shows, CPS \cite{chen2021semi} and MeanTeacher \cite{tarvainen2017mean}  ignore some tooth morphology and did not obtain complete segmentation results.  FixMatch \cite{sohn2020fixmatch}  is not accurate enough to segment the edges of the teeth. Instead, regardless of tooth morphology or edge  information, our method achieves segmentation that is superior to comparison methods.

\begin{table*}[h]
	\centering
	\caption{The quantitative evaluation results  of our method and the comparison methods on the validation set. }
	\label{tab:table_val}
	\begin{tabularx}{\textwidth}{XXXXX} \hline
		Method & Score  & Dice & IoU & H(d) \\  		\hline
		CPS \cite{chen2021semi}  & 0.8582  & 0.8501 & 0.8078 & 0.0808 \\ 
  		FixMatch \cite{sohn2020fixmatch} & 0.8796  & 0.8713 & 0.8266 & 0.0563 \\ 
	MeanTeacher~\cite{tarvainen2017mean} & 0.8737  & 0.8645 & 0.8210 & 0.0613 \\ 
 	Ours & 0.8859  & 0.8801 & 0.8425 & 0.0632\\ 
		\hline
	\end{tabularx}
\end{table*}

The quantitative evaluation results are summarized in \tabref{tab:table_val}. As \tabref{tab:table_val} shows, our proposed method taking advantage of its multi-stage training and domain adaptation mechanism achieves a mean Dice of 0.8801, a mean IoU of  0.8425 and a mean Score of 0.8859, which exceeds the comparison methods.

\section{Conclusion}
In this paper,  we propose a multi-stage framework for 3D  tooth segmentation, which achieves the third place in the STS-3D challenge. The experiments on validation set compared with other semi-supervised segmentation methods further indicate the validity of our approach.

In the future work, in order to better utilize the information of the labeled data, the appropriate introduction of perturbations at the image level or the feature level is considered, and the internal knowledge migration using the predictive consistency of the model under different strengths of perturbations is carried out by using the knowledge distillation method, so as to optimize the learning effectiveness and generalization ability of the model on unlabeled data.

\section{Authors’ Contributions}
Chunshi Wang and Bin Zhao contribute equally to this paper. 

\subsubsection*{Acknowledgments.} This work is supported in part by the National Natural Science Foundation of China (Grant No.62076077),  the Project of Improving the Basic Scientific Research Ability of Young and Middle-Aged Teachers in Universities of Guangxi Province (Grant No.2023KY0223), Youth Science Foundation of Guangxi Natural Science Foundation (Grant No.2023GXNSFBA026018) and the Guangxi Science and Technology Major Project (Grant No.AA22068057).

\bibliographystyle{unsrt}

\bibliography{reference}

\begin{thebibliography}{10}

\bibitem{jang2021fully}
Tae~Jun Jang, Kang~Cheol Kim, Hyun~Cheol Cho, and Jin~Keun Seo.
\newblock A fully automated method for 3d individual tooth identification and segmentation in dental cbct.
\newblock {\em IEEE transactions on pattern analysis and machine intelligence}, 44(10):6562--6568, 2021.

\bibitem{gao2010individual}
Hui Gao and Oksam Chae.
\newblock Individual tooth segmentation from ct images using level set method with shape and intensity prior.
\newblock {\em Pattern Recognition}, 43(7):2406--2417, 2010.

\bibitem{ji2014level}
Dong~Xu Ji, Sim~Heng Ong, and Kelvin Weng~Chiong Foong.
\newblock A level-set based approach for anterior teeth segmentation in cone beam computed tomography images.
\newblock {\em Computers in biology and medicine}, 50:116--128, 2014.

\bibitem{hiew2010tooth}
LT~Hiew, SH~Ong, Kelvin~WC Foong, and C~Weng.
\newblock Tooth segmentation from cone-beam ct using graph cut.
\newblock In {\em Proceedings of the Second APSIPA Annual Summit and Conference}, pages 272--275, 2010.

\bibitem{evain2017semi}
Timoth{\'e}e Evain, Xavier Ripoche, Jamal Atif, and Isabelle Bloch.
\newblock Semi-automatic teeth segmentation in cone-beam computed tomography by graph-cut with statistical shape priors.
\newblock In {\em 2017 IEEE 14th International Symposium on Biomedical Imaging (ISBI 2017)}, pages 1197--1200, 2017.

\bibitem{polizzi2023tooth}
Alessandro Polizzi, Vincenzo Quinzi, Vincenzo Ronsivalle, Pietro Venezia, Simona Santonocito, Antonino Lo~Giudice, Rosalia Leonardi, and Gaetano Isola.
\newblock Tooth automatic segmentation from cbct images: a systematic review.
\newblock {\em Clinical Oral Investigations}, pages 1--16, 2023.

\bibitem{chen2020automatic}
Yanlin Chen, Haiyan Du, Zhaoqiang Yun, Shuo Yang, Zhenhui Dai, Liming Zhong, Qianjin Feng, and Wei Yang.
\newblock Automatic segmentation of individual tooth in dental cbct images from tooth surface map by a multi-task fcn.
\newblock {\em IEEE Access}, 8:97296--97309, 2020.

\bibitem{hsu2022improving}
Kang Hsu, Da-Yo Yuh, Shao-Chieh Lin, Pin-Sian Lyu, Guan-Xin Pan, Yi-Chun Zhuang, Chia-Ching Chang, Hsu-Hsia Peng, Tung-Yang Lee, Cheng-Hsuan Juan, et~al.
\newblock Improving performance of deep learning models using 3.5 d u-net via majority voting for tooth segmentation on cone beam computed tomography.
\newblock {\em Scientific Reports}, 12(1):19809, 2022.

\bibitem{cui2019toothnet}
Zhiming Cui, Changjian Li, and Wenping Wang.
\newblock Toothnet: automatic tooth instance segmentation and identification from cone beam ct images.
\newblock In {\em Proceedings of the IEEE/CVF conference on computer vision and pattern recognition}, pages 6368--6377, 2019.

\bibitem{cui2022fully}
Zhiming Cui, Yu~Fang, Lanzhuju Mei, Bojun Zhang, Bo~Yu, Jiameng Liu, Caiwen Jiang, Yuhang Sun, Lei Ma, Jiawei Huang, et~al.
\newblock A fully automatic ai system for tooth and alveolar bone segmentation from cone-beam ct images.
\newblock {\em Nature communications}, 13(1):2096, 2022.

\bibitem{cui2022ctooth}
Weiwei Cui, Yaqi Wang, Qianni Zhang, Huiyu Zhou, Dan Song, Xingyong Zuo, Gangyong Jia, and Liaoyuan Zeng.
\newblock Ctooth: a fully annotated 3d dataset and benchmark for tooth volume segmentation on cone beam computed tomography images.
\newblock In {\em International Conference on Intelligent Robotics and Applications}, pages 191--200, 2022.

\bibitem{zhao14automatic}
Bin Zhao, Shuxue Ding, Hong Wu, Guohua Liu, Chen Cao, Song Jin, and Zhiyang Liu.
\newblock Automatic acute ischemic stroke lesion segmentation using semi-supervised learning.
\newblock {\em International Journal of Computational Intelligence Systems}, 14(1):723--733, 2021.

\bibitem{zhao2022combine}
Bin Zhao, Zhiyang Liu, Guohua Liu, Mengran Wu, Chen Cao, Song Jin, Hong Wu, and Shuxue Ding.
\newblock Combine unlabeled with labeled mr images to measure acute ischemic stroke lesion by stepwise learning.
\newblock {\em IET Image Processing}, 16(14):3965--3976, 2022.

\bibitem{cui2022ctooth+}
Weiwei Cui, Yaqi Wang, Yilong Li, Dan Song, Xingyong Zuo, Jiaojiao Wang, Yifan Zhang, Huiyu Zhou, Bung~san Chong, Liaoyuan Zeng, et~al.
\newblock Ctooth+: A large-scale dental cone beam computed tomography dataset and benchmark for tooth volume segmentation.
\newblock In {\em MICCAI Workshop on Data Augmentation, Labelling, and Imperfections}, pages 64--73, 2022.

\bibitem{isensee2021nnu}
Fabian Isensee, Paul~F Jaeger, Simon~AA Kohl, Jens Petersen, and Klaus~H Maier-Hein.
\newblock nnu-net: a self-configuring method for deep learning-based biomedical image segmentation.
\newblock {\em Nature methods}, 18(2):203--211, 2021.

\bibitem{yao2022enhancing}
Huifeng Yao, Xiaowei Hu, and Xiaomeng Li.
\newblock Enhancing pseudo label quality for semi-supervised domain-generalized medical image segmentation.
\newblock In {\em Proceedings of the AAAI Conference on Artificial Intelligence}, volume~36, pages 3099--3107, 2022.

\bibitem{wang2022freematch}
Yidong Wang, Hao Chen, Qiang Heng, Wenxin Hou, Yue Fan, Zhen Wu, Jindong Wang, Marios Savvides, Takahiro Shinozaki, Bhiksha Raj, et~al.
\newblock Freematch: Self-adaptive thresholding for semi-supervised learning.
\newblock In {\em The Eleventh International Conference on Learning Representations}, 2022.

\bibitem{yang2023revisiting}
Lihe Yang, Lei Qi, Litong Feng, Wayne Zhang, and Yinghuan Shi.
\newblock Revisiting weak-to-strong consistency in semi-supervised semantic segmentation.
\newblock In {\em Proceedings of the IEEE/CVF Conference on Computer Vision and Pattern Recognition}, pages 7236--7246, 2023.

\bibitem{klambauer2017self}
G{\"u}nter Klambauer, Thomas Unterthiner, Andreas Mayr, and Sepp Hochreiter.
\newblock Self-normalizing neural networks.
\newblock {\em Advances in neural information processing systems}, 30, 2017.

\bibitem{liu2018towards}
Zhiyang Liu, Chen Cao, Shuxue Ding, Zhiang Liu, Tong Han, and Sheng Liu.
\newblock Towards clinical diagnosis: Automated stroke lesion segmentation on multi-spectral mr image using convolutional neural network.
\newblock {\em IEEE Access}, 6:57006--57016, 2018.

\bibitem{loshchilov2017decoupled}
Ilya Loshchilov and Frank Hutter.
\newblock Decoupled weight decay regularization.
\newblock {\em arXiv preprint arXiv:1711.05101}, 2017.

\bibitem{chen2021semi}
Xiaokang Chen, Yuhui Yuan, Gang Zeng, and Jingdong Wang.
\newblock Semi-supervised semantic segmentation with cross pseudo supervision.
\newblock In {\em Proceedings of the IEEE/CVF Conference on Computer Vision and Pattern Recognition}, pages 2613--2622, 2021.

\bibitem{sohn2020fixmatch}
Kihyuk Sohn, David Berthelot, Nicholas Carlini, Zizhao Zhang, Han Zhang, Colin~A Raffel, Ekin~Dogus Cubuk, Alexey Kurakin, and Chun-Liang Li.
\newblock Fixmatch: Simplifying semi-supervised learning with consistency and confidence.
\newblock {\em Advances in neural information processing systems}, 33:596--608, 2020.

\bibitem{tarvainen2017mean}
Antti Tarvainen and Harri Valpola.
\newblock Mean teachers are better role models: Weight-averaged consistency targets improve semi-supervised deep learning results.
\newblock {\em Advances in neural information processing systems}, 30, 2017.

\end{thebibliography}

\end{document}